%% file: main.tex
\documentclass{ecai} 

\usepackage{latexsym}
\usepackage{amssymb}
\usepackage{amsmath}
\usepackage{amsthm}
\usepackage{booktabs}
\usepackage{enumitem}
\usepackage{graphicx}
\usepackage{color}
\usepackage{overpic}

\usepackage{tikz}
\usepackage{pgfplots}
\usepackage{xcolor}
\usepackage{subfigure}
\usepackage{filecontents}
\usepackage{multirow}

\usepackage{listings}
\usepackage{float}
\usepackage{amsmath}
\usepackage{bbding}
\usepackage{rotating}
\usepackage{pifont}
\usepackage{tikz,pgfplots}

\usepackage{booktabs}       
\usepackage{amsfonts}       
\usepackage{nicefrac}       
\usepackage{microtype}      
\usepackage{xcolor}         
\usepackage{colortbl}
\usepackage{overpic}
\usepackage{indentfirst}
\usepackage{latexsym}
\usepackage{times}
\usepackage{soul}
\usepackage[utf8]{inputenc}
\usepackage[small]{caption}
\usepackage{booktabs}
\usepackage[switch]{lineno}
\usepackage{amsmath}
\usepackage{amssymb}
\usepackage{amsthm}
\usepackage{mathrsfs}
\usepackage{amsfonts}
\usepackage{float}
\usepackage{multirow}
\usepackage{booktabs}
\usepackage{amsthm}
\usepackage{xspace}
\usepackage{subfig}
\usepackage{booktabs}
\usepackage{color}
\usepackage{xcolor}
\usepackage{tabularx}
\usepackage{graphicx}
\usepackage{amsmath}
\usepackage{amssymb}
\usepackage{amsthm}
\usepackage{mathrsfs}
\usepackage{algorithm}
\usepackage{algpseudocode}
\usepackage{amsfonts}
\usepackage{float}

\usepackage[capitalize]{cleveref}
\crefname{section}{Sec.}{Secs.}
\Crefname{section}{Section}{Sections}
\Crefname{table}{Table}{Tables}
\crefname{table}{Tab.}{Tabs.}

\newcommand{\BibTeX}{B\kern-.05em{\sc i\kern-.025em b}\kern-.08em\TeX}

\begin{document}


\begin{frontmatter}


\title{Fully Fine-tuned CLIP Models are Efficient\\ Few-Shot Learners}

\begin{abstract}

\author{
   Mushui Liu,
   Bozheng Li,
   Yunlong Yu \\
   Zhejiang University
}

Prompt tuning, which involves training a small set of parameters, effectively enhances the pre-trained Vision-Language Models (VLMs) to downstream tasks. However, they often come at the cost of flexibility and adaptability when the tuned models are applied to different datasets or domains. In this paper, we explore capturing the task-specific information via meticulous refinement of entire VLMs, with minimal parameter adjustments. When fine-tuning the entire VLMs for specific tasks under limited supervision, overfitting and catastrophic forgetting become the defacto factors. To mitigate these issues, we propose a framework named CLIP-CITE via designing a discriminative visual-text task, further aligning the visual-text semantics in a supervision manner, and integrating knowledge distillation techniques to preserve the gained knowledge. Extensive experimental results under few-shot learning, base-to-new generalization, domain generalization, and cross-domain generalization settings, demonstrate that our method effectively enhances the performance on specific tasks under limited supervision while preserving the versatility of the VLMs on other datasets.
\end{abstract}

\end{frontmatter}


\input{tex/intro}
\input{tex/related}

\input{tex/method}

\input{tex/exp}

\input{tex/conclusion}

\bibliography{main}

\end{document}

%% file: tex/intro.tex
\section{Introduction}
Recently, the pre-trained Vision-Language Models (VLMs) such as CLIP \cite{clip} and ALIGN \cite{ALIGN} have demonstrated impressive generalization capabilities across various downstream tasks, including image recognition \cite{coop,cocoop}, object detection \cite{gu2021open}, image segmentation \cite{denseclip}, and action recognition \cite{rasheed2023fine}. Though versatile, the performance of the VLMs on specific domains shows considerable potential for improvement, especially under limited supervision \cite{coop}. 

\begin{figure}[!htb]
    \centering
     \setlength{\tabcolsep}{4pt}
     \begin{tabular}{cc}   
     \multicolumn{2}{c}{
     \includegraphics[width=1\linewidth]{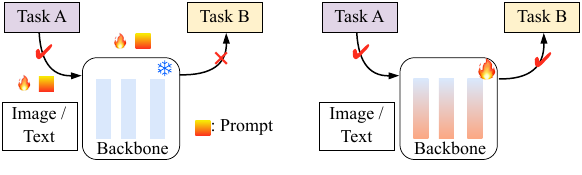}}
    \\  
    ~~~~\footnotesize (a) Prompt-based Tuning & \footnotesize ~~(b) Fine-tuning
    \vspace{8pt}
    \\ 
    \multicolumn{2}{c}{
        \small
       \begin{tabular}{l@{\hspace{0.5cm}}|cc@{\hspace{0.3cm}}|@{\hspace{0.3cm}}cc}
        \toprule
        \multirow{2}{*}{Method} & \multicolumn{4}{c}{\textbf{EuroSAT} $\Rightarrow$ \textbf{ImageNet}} \\
        & \multicolumn{2}{c}{\textbf{Base} $\Rightarrow$ \textbf{New}} & \textbf{Base} & \textbf{New} \\
        \midrule
        CLIP \cite{clip} & 56.48 & 64.05 & \textbf{72.43} & 68.14 \\
        CoOp \cite{coop} & 92.19 & 54.74 & 51.81 & 49.69 \\
        MaPLe \cite{maple} & 95.31 & 71.28 & 36.88 & 44.76 \\
        \midrule
        FT-Probe \cite{clip} & 60.86 & 71.34 & 72.03 & 68.04 \\
        CLIP-CITE (Ours) & \textbf{95.61} & \textbf{80.59} & 72.29 & \textbf{68.38} \\
        \bottomrule
        \end{tabular}
    }
    \vspace{1mm}
    \\
    \multicolumn{2}{c}{
       \footnotesize (c) Cross-Domain Generalization Experiments
    } \\
     \end{tabular}
    \vspace{10pt}
    \caption{(a) Prompt-based methods introduce a few trainable prompts to incorporate task-specific knowledge. (b) Fine-tuning methods adjust the whole model to adapt to the specific tasks. (c) Comparison results (\%) under the cross-domain generalization setting in the limited-data regime.}
     \vspace{10pt}
    \label{fig:ft-methods}
\end{figure}


The existing methods attempt to equip VLMs with domain-specific knowledge by employing various tuning techniques. Prompt-based Tuning \cite{coop,cocoop,maple,vpt} (as shown in \cref{fig:ft-methods}~(a)), which refines the pre-trained models with specific prompts while keeping the parameters of VLMs fixed, has gained popularity due to its efficient parameter utilization and capability of quickly adapting VLMs to domain-specific information. 

While the prompt-based tuning strategies enable VLMs to effectively capture domain-specific information with limited supervision \cite{coop,cocoop,maple,vpt}, there is a risk that these strategies may compromise the versatility of VLMs. In other words, the prompts trained on domain-specific data may struggle to generalize to other domains, limiting their versatility. A piece of evidence is provided in \cref{fig:ft-methods}(c), which shows a transfer experiment under the cross-domain generalization setting. We employ a few-shot setting to train the model using the EuroSAT base training set, followed by evaluating its performance on both EuroSAT and ImageNet datasets. While the prompt-based methods, i.e., CoOp \cite{coop} and MaPLe \cite{maple} significantly improve the EuroSAT dataset results, they are at the cost of sacrificing their generalizability on the other datasets. In particular, their performances on the ImageNet dataset severely lag behind those of the zero-shot CLIP model.

In this work, we restate the \textbf{professionalism} and \textbf{versatility} of VLMs. Professionalism highlights the ability of VLMs to excel in specific domains, categories, and tasks, while versatility highlights their capability to perform across various domains, categories, and tasks. Based on the previous analysis, the prompt-based approaches improve the VLMs' professionalism but compromise their versatility.



When customizing VLMs to specific domains, fine-tuning the entire models would distribute task-specific knowledge across all parameters (as illustrated in \cref{fig:ft-methods}.~(b)). Unlike prompt-based tuning techniques, strategies that involve fine-tuning the entire VLMs have been relatively under-explored and under-appreciated, particularly in limited data regimes, due to the significant number of training parameters involved. One example is the FT-Probe \cite{clip}, which employs a straightforward strategy of fine-tuning the entire model and incorporating a linear probe on top of the visual representations. This approach allows the model to preserve the model's versatility during the adaptation for specific domains, as evident from the results achieved by FT-Probe, presented in \cref{fig:ft-methods}(c). However, this fine-tuning strategy has demonstrated only marginal improvements in specific domains compared to prompt-based competitors. We posit that the limited supervision when tailoring the models leads to the emergence of an overfitting issue, which undermines the fine-tuning strategy's effectiveness in specific domains. More evidence is provided in supplementary materials.

In this paper, we propose a fine-tuning method called \textbf{CLIP-CITE} that enhances the CLIP's professionalism on specific domains while preserving its versatility by primarily enhan\textbf{C}ing the capability of the \textbf{I}mage-\textbf{T}ext alignm\textbf{E}nt task. Specifically, our CLIP-CITE approach incorporates three key aspects. Firstly, to quickly equip the domain-specific information for CLIP, our CLIP-CITE connects the alignment score with the classification probability in a way that prioritizes higher alignment scores for image-text pairs belonging to the same class. Secondly, our approach fine-tunes the entire model using an image-text alignment task, aligning with the original training objective of the pre-trained CLIP model. This differs from the classification task utilized in \cite{clip}, ensuring a consistent training objective throughout the adaptation process. Note that training an image-text alignment task usually requires a large batch \cite{clip,flyp} in implementation, posing a significant challenge when working with limited data regimes. To overcome this issue, we propose utilizing a class-level image-text alignment task as an alternative to the original instance-level alignment task. Finally, to alleviate the catastrophic forgetting issue, we introduce a vision-language similarity distillation strategy. This strategy regularizes the model by transferring the image-text alignment relationship learned by the pre-trained CLIP model, further ensuring a minimal change in parameters. As shown in the last row of \cref{fig:ft-methods}~(c), our CLIP-CITE enhances EuroSAT dataset performance while simultaneously upholding generalization capability on the ImageNet dataset. 


In summary, our highlights are as follows:
\begin{itemize}
   \item We propose CLIP-CITE, a simple but efficient fine-tuning method that enhances the VLMs' professionalism while maintaining their versatility under limited data supervision. CLIP-CITE comprehensively fine-tunes CLIP to enable it to promptly incorporate task-specific information through enhanced image-text alignment and safeguard the learned knowledge. 
   

    \item We evaluate CLIP-CITE through experiments in different settings, including few-shot image recognition, base-to-new generalization, domain generalization, and cross-domain scenarios. The experimental results demonstrate that CLIP-CITE not only sets new benchmarks in these tasks on specific datasets, but also preserves the original versatility of CLIP on other datasets.
\end{itemize}



%% file: tex/related.tex
\section{Related work}

\subsection{Vision-Language Model} 
Recent years have witnessed remarkable achievements on large-scale pre-trained vision-language models \cite{clip,ALIGN,wang2021simvlm,alayrac2022flamingo,wang2022beit,huang2023clover}. Representatively, CLIP, ALIGN \cite{clip,ALIGN} jointly associate the images and their corresponding text descriptions by optimizing a contrastive objective. Training on the millions of image-text pairs, CLIP aligns the image and language space, showing the powerful generalization on downstream tasks. Based on CLIP, many works seek to transfer the model to special tasks, e.g., few-shot image recognition \cite{coop,cocoop,maple}, segmentation \cite{denseclip}, and action recognition \cite{rasheed2023fine}. In this paper, we also leverage the benefits of multi-modal alignment and the generalization ability of CLIP. By fine-tuning the CLIP model in limited data regimes, we investigate how the model can adapt its knowledge and generalize to perform well in this particular challenging scenario.

\subsection{Few-Shot Transfer Learning Based on CLIP}
Prompt tuning \cite{coop,cocoop,vpt,maple} and fine-tuning \cite{clipood,wiseft,flyp,lp-ft} are two main methods to transfer the CLIP to the downstream tasks. Prompt tuning is widely used in language models \cite{houlsby2019parameter,liu2023pre}, which raises attention in vision and multi-modality areas \cite{coop,vpt,sam}. Context Optimization (CoOp) \cite{coop} improves the downstream few-shot image recognition tasks via learning the soft textual prompts. CoCoOp \cite{cocoop} and MaPLe \cite{maple} further boost the generalization ability through the image-condition information and multi-modal prompts, respectively. Except for the textural prompts, Visual Prompt Tuning (VPT) \cite{vpt} introduces the vision prompts on the large vision models. Although these prompt tuning methods show efficient and excellent performance, they may fail to overfit the task-specific distribution. 


As the alternative, fine-tuning methods directly optimize the model under task-specific situations. WiSE-FT \cite{wiseft}, LP-FT \cite{lp-ft} achieves the robustness of fine-tuning via a weight-ensemble manner. CLIPood \cite{clipood} further finetunes the model via the text semantic similarity and model ensemble under an out-of-distribution situation. A similar work related to our method is FLYP \cite{flyp}, which fine-tunes the CLIP model via the pre-trained contrastive objective to obtain the multi-modal alignment ability. In comparison, our method distinguishes the supervised vision-language pairs and incorporates the task-specific into the fine-tuning process. Leveraging this improved image-text alignment task, our method aims to perform more robustly under limited supervision.

\begin{figure*}[!ht]
    \centering
    \includegraphics[width=\linewidth]{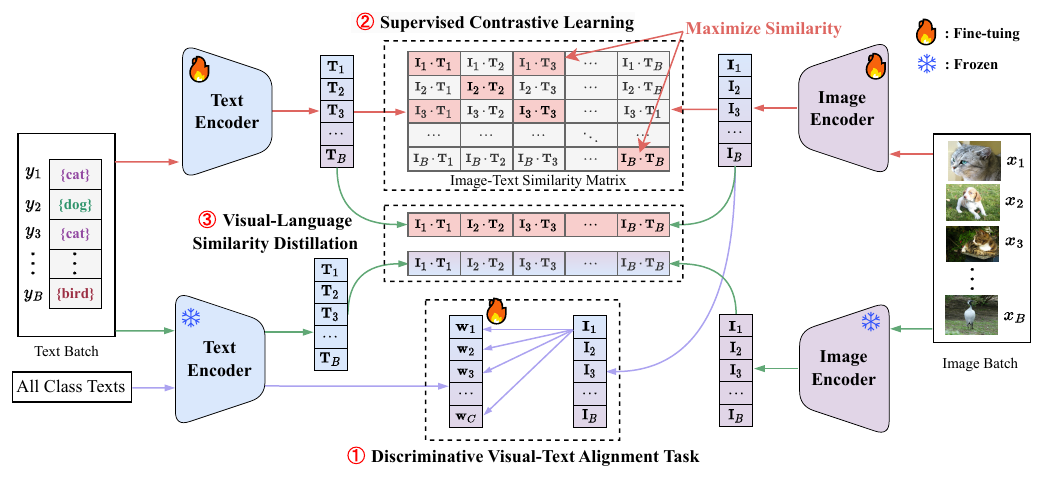}
    \vspace{10pt}
    \caption{The framework of our CLIP-CITE method. CLIP-CITE fine-tunes the whole CLIP model with a \textcolor{red}{\ding{172}} discriminative visual-text alignment task and a \textcolor{red}{\ding{173}} supervised contrastive loss to enhance the image-text alignment in downstream tasks. Moreover, a \textcolor{red}{\ding{174}} vision-language similarity distillation loss incorporates the generalization knowledge of the pre-trained CLIP model into the fine-tuned model.}
    \vspace{5pt}
    \label{fig:framework}
\end{figure*}

%% file: tex/method.tex
\section{Method}

In this work, we fine-tune the CLIP models \cite{clip} for the scenarios with limited data available. The architecture of CLIP includes two key components: a visual encoder denoted as $\theta_{I}$ and a text encoder denoted as $\theta_{T}$. By aligning language and visual modalities on 400 million text-image web data, CLIP is endowed with zero-shot and open-vocabulary capabilities. 

To perform zero-shot classification, CLIP utilizes handcrafted text prompts with class labels. These prompts consist of a predefined set of class labels denoted as $y \in \{y_1, y_2, ..., y_C\}$, where $C$ represents the total number of classes. Each prompt typically takes the form of ``a photo of a [category]", where ``[category]" corresponds to the class label name. Then, the image prediction $\hat{y}$ corresponding to the class $i$ is obtained by calculating the cosine similarity scores between the image embedding $\mathbf{I}$ and the text embedding $\mathbf{T}$, which is formulated as: 
\begin{equation}
p(\hat{y}|x)=\frac{\exp \left(s \left(\mathbf{I}, \mathbf{T}_i\right) / \tau\right)}{\sum_{c=1}^C \exp \left(s \left(\mathbf{I}, \mathbf{T}_i \right)/ \tau \right)},
\label{eq:zs_clip}
\end{equation}
where $s( \cdot )$ is the similarity metric, $\tau$ denotes the temperature parameter. By calculating the softmax probabilities using the similarity scores, CLIP can assign a class label to the image, even if it has not been explicitly trained on that specific class. 

Although CLIP has demonstrated impressive zero-shot performance, its integration into specific downstream tasks still requires further refinements through subtle adjustments. Extensive prompt-based methods \cite{coop,cocoop,maple} have been proposed to enhance CLIP's performance in specific contexts. In this study, we investigate the underestimated fine-tuning strategy and propose to improve the fine-tuning method from the perspectives of task designing, multi-modal alignment, and knowledge preservation. As illustrated in \cref{fig:framework}, our framework comprises three components, i.e., discriminative visual-text alignment task, supersized contrastive learning, and vision-language similarity distillation.

\subsection{Discriminative Visual-Text Alignment task}
Naive fine-tuning methods for downstream classification tasks typically involve adding a randomly initialized linear classifier on top of the pre-trained visual encoder \cite{clip,lp-ft}. The whole model is then fine-tuned using the available domain-specific data for the classification task at hand. However, this training strategy often leads to overfitting on the limited available training data, resulting in poor generalization performance on unseen data.

To address this limitation, we propose to fine-tune the model with a discriminative visual-text alignment task that combines visual-semantic alignment and image classification. Specifically, we connect the similarity scores between the visual and the text embeddings with the probability that the visual image belongs to the class associated with the text embedding, which is formulated:
\begin{align}
p(\hat{y}|x)=\frac{\exp \left(s \left(\theta_{I}\left(x\right), \theta_{T}\left(t_i\right)\right)\right)}{\sum_{c=1}^C \exp \left(s \left(\theta_{I}\left(x\right), \theta_{T}\left(t_c\right)\right) \right)},
\label{eq:combine}
\end{align}
where $s(\cdot)$ is the consine similarity, $\theta_{I}$ and $\theta_{T}$ denotes the visual encoder and text encoder, respectively, $t_i$ is the text description of class $i$, which is obtained in the form of ``a photo of a [category]", where ``[category]" corresponds to one of the class labels. 

Note that \cref{eq:combine} is equivalent to initializing the parameters of the visual classifier ${W}~=~\{w_i\}_{i=0}^{C},~w_i~=~\theta_{T}\left(t_i\right)$ with the embeddings of the text descriptions of all the available classes and is consistent with the prediction of the test data.   To this end, the objective loss of the discriminative visual-text alignment task is:
\begin{equation}
    \label{eq:dva}
    \mathcal{L}_{DVA}= -\sum_{x\in \mathcal{B}} \log p(\hat{y}|x),
\end{equation}
where $\mathcal{B}$ denotes a training batch during the fine-tuning process. To quickly adapt the model to the target classification task, we freeze the text encoder and take $W$ and $\theta_{I}$ as the learnable parameter to fine-tune. Through fine-tuning this task, the model acquires the ability to collaboratively associate visual and textual representations, thereby enhancing its capacity to utilize semantic information effectively for the discriminative task.

\subsection{Supervised Contrastive Learning}



To preserve and enhance the representation capability of the pre-trained CLIP, we argue that aligning image and text remains essential, as it corresponds to the task employed in the training of the original CLIP models. However, it is worth noting that aligning images and texts can often require a large batch size, which may not be suitable in situations where data availability is limited.

To mitigate this limitation, we customize an image-text alignment strategy to fine-tune the whole CLIP models (including both $\theta_{I}$ and $\theta_{T}$) under the limited data regimes. Specifically, we adopt a supervised contrastive loss to align images and texts. Given a pair of data $(x, t)$, where $t$ is derived from the category of $x$ in the form of ``a photo of a [category]", the supervised contrastive loss is defined as:
\begin{align} \label{eq:itm}
\mathcal{L}_{SCL} &= \sum_{x_i \in \mathcal{B}} \log \frac{\exp \left(s \left(\theta_{I}\left(x_i\right), \theta_{T}\left(t_i\right)\right)\right)}{\sum_{t_j\in B }  \mathbb{I}_{t_j \ne x_i} \cdot \exp \left(s \left(\theta_{I}\left(x_i\right), \theta_{T}\left(t_j\right)\right)\right)} \\ \nonumber
& + \sum_{t_i \in \mathcal{B}} \log \frac{\exp \left(s \left( \theta_{T}\left(t_i\right)\right), \theta_{I}\left(x_i\right)\right)}{\sum_{x_j\in B} \mathbb{I}_{x_j \ne t_i} \cdot \exp \left(s \left( \theta_{T}\left(t_i\right)\right), \theta_{I}\left(x_j\right) \right)}, 
\end{align}
where $s$ denotes the cosine similarity, $\mathcal{B}$ denotes a training batch, and $\mathbb{I}$ denotes the category indicator function. Notably, $\mathcal{L}_{SCL}$ can be considered a special case of FLYP \cite{flyp} in scenarios where there are no same class instances within the batch, employing unsupervised contrastive loss to optimize image-text alignment. 


The designed supervised contrastive loss encourages the model to learn representations that bring similar images and their associated text embeddings closer together while pushing apart images and their non-matching text embeddings. By enforcing this alignment, the model can better capture the semantic relationship between images and their associated text while preserving and enhancing the representation capability of the pre-trained CLIP in specific domains.

\subsection{Vision-Language Similarity Distillation}
While fine-tuning can improve performance on downstream tasks, it would suffer from potential challenges such as catastrophic forgetting and decreased generalization capabilities on the other datasets. To remedy this issue, we introduce a novel vision-language similarity distillation loss to distill the modal consistency from the pre-trained CLIP to the fine-tuned model. Specifically, the vision-language similarity distillation loss is defined as:

\begin{equation}
    \label{eq:vls}
    \mathcal{L}_{VLD} = \sum_{x \in \mathcal{B}} \mathcal{D}_{KL} 
    \left(p\left(\hat{y}|x\right), 
          \hat{p}\left(\hat{y}|x\right)
    \right), 
\end{equation}
where $p\left(\hat{y}|x\right)$, derived from \cref{eq:zs_clip}, is computed using fine-tuned models $\theta_{I}$ and $\theta_{T}$ to determine batch cosine image-text similarity scores. While $\hat{p}\left(\hat{y}|x\right)$, also obtained by \cref{eq:zs_clip}, applies the original CLIP models. $\mathcal{D}_{KL}$ denotes the Kullback-Leibler divergence. Note that the batch cosine image-text similarity scores undergo normalization through a softmax function to establish a probability distribution.

By minimizing the Kullback-Leibler divergence between the distributions of image-text similarity calculated from the original CLIP encoders and those from the fine-tuned encoders, CLIP-CITE encourages the fine-tuned model to acquire comparable modal alignments and image-text relationship within batch as the pre-trained CLIP models. This strategy upholds modal consistency and facilitates the transfer of knowledge from the pre-trained model to the fine-tuned model.




\subsection{Final Objective Function}
To fine-tune the whole CLIP models, we combine \cref{eq:dva}, \cref{eq:itm}, and \cref{eq:vls}, obtaining the final objective loss:
\begin{equation}
    \label{eq:total-loss}
    \mathcal{L} = \mathcal{L}_{DVA} + \lambda \cdot \mathcal{L}_{SCL} + \eta \cdot \mathcal{L}_{VLD},
\end{equation}
where $\lambda$ and $\eta$ are the two hypermeters to balance the items. After the fine-tuning process, we obtain the updated visual encoder $\theta_{I}$ and text encoder $\theta_{T}$. 

During inference, we use a weighted ensemble proposed by \cite{wiseft} to combine the fine-tuned model and the pre-trained model:
\begin{equation}
\label{eq: weight ensemble}
    \hat{\theta}_{I} = \alpha \cdot \theta_{I} + (1 - \alpha) \cdot \theta_{I}^{zs}, \quad \hat{\theta}_{T} = \alpha \cdot \theta_{T} + (1 - \alpha) \cdot \theta_{T}^{zs},
\end{equation}
where $\alpha$ is a hyperparameter. Different from \cite{wiseft} that only considers ensemble in the visual modality, the text encoder in our method is optimized during the fine-tuning process, so the text modality is further considered in this work.

%% file: tex/exp.tex
\section{Experiments}

\subsection{Experiment Settings}

To assess the efficacy of our method, we conduct experiments within the few-shot learning paradigm. In this setup, the model undergoes training using some base classes, each of which is represented by a limited number of samples. Subsequently, the model's performance is evaluated on the novel classes. Based on the origin of the classes and the domains to which the base and novel data belong, the evaluated tasks are categorized into four distinct groups: few-shot learning (\textbf{FSL}), domain generalization (\textbf{DG}), base-to-new generalization (\textbf{BNG}), and cross-domain generalization (\textbf{CDG}).

\textbf{FSL.} In FSL, the training data and test data are from the same classes and the same domain, which assesses the model's effectiveness in the limited supervision scenario. 


\textbf{DG.} In DG, the training data and test data are from the same classes but in different domains. 



\textbf{BNG.} In BNG, the training data and test data are from different classes but in the same domain, which evaluates the model's ability to generalize to new and previously unseen classes, thereby gauging its open-vocabulary generalization capability. 

\textbf{CDG.} In CDG, the training data and test data are from different classes and different domains. In the experiments, the model is trained on the base classes of dataset A and evaluated on the new classes of dataset B.




\textbf{Dataset Settings.} For FSL, BNG, and CDG settings, we use 11 image classification datasets, i.e., ImageNet \cite{ImageNet} and Caltech-101 \cite{caltech-101} for generic object classification; OxfordPets \cite{OxfordPets}, StanfordCars \cite{cars}, Flowers \cite{Flowers}, Food101 \cite{food101}, and FGVCAircraft \cite{FGVCAircraft} for fine-grained visual categorization, EuroSAT \cite{eurosat} for satellite image classification, UCF101 \cite{ucf101} for action recognition, DTD \cite{dtd} for texture classification, and SUN397 \cite{sun397} for scene recognition. We randomly sample 16 images (shots) from each class in all the datasets mentioned above in BNG scinario. For the DG, we treat the ImageNet as the source domain, and the ImageNetV2 \cite{ImageNetV2}, ImageNet-Sketch \cite{imagenet-sketch}, ImageNet-A \cite{imagenet-a} and ImageNet-R \cite{imagenet-r} as the target domains for evaluation. 

\textbf{Implement Details.} In our implementation, we leverage the pre-trained ViT-B/16 model from CLIP \cite{clip} for evaluation purposes. We employ the AdamW optimizer, incorporating the cosine annealing strategy to fine-tune our model. The initial learning rate is fixed at 5e-6, while the batch size is set to 32 for most datasets. However, for the EuroSAT dataset, we use a batch size of 16, and for ImageNet, it is increased to 64. The hyperparameters $\lambda$, $\eta$, and $\alpha$ are consistently set to 0.7, 0.1, and 0.5 across all experiments, respectively. We train our model for 20 epochs. All input images are randomly resized and cropped to a resolution of 224 × 224 pixels. No additional data augmentation techniques are employed, apart from random resizing and cropping. For reproducibility, we report the average results of CLIP-CITE across three distinct random seeds for each experiment.


\begin{figure}[t]
    \centering
     \includegraphics[width=0.95\linewidth]{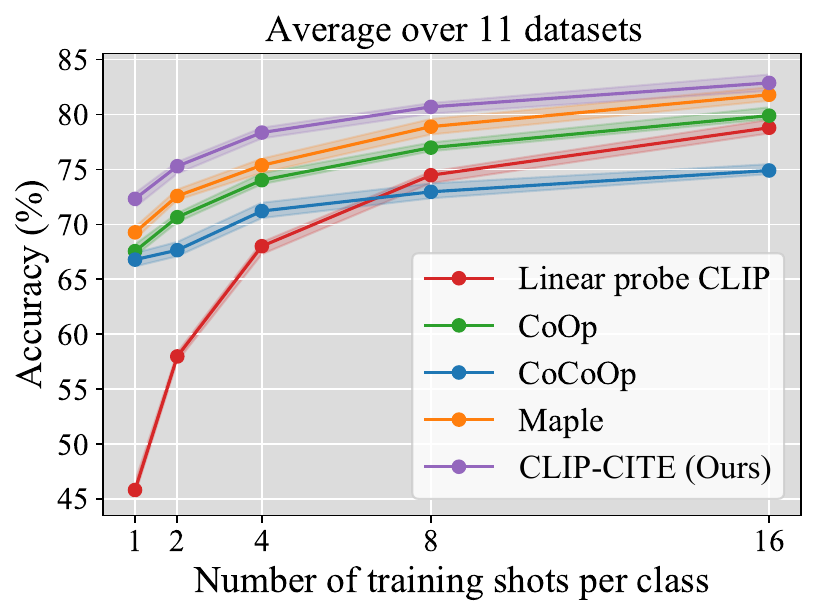}    
     \vspace{6pt}

    \caption{FSL Comparison results of our CILP-CITE and four competitors on the 11 datasets. All of the methods are trained on the ViT-B/16 backbone and implemented with the same experimental settings. We report the average performance of 11 datasets.}
    \vspace{10pt}
   \label{fig:fsl}
\end{figure}

\subsection{Performance Comparison}
\textbf{Results of FSL.}  \cref{fig:fsl} presents the average results of four competitors and our CLIP-CITE on the 11 datasets under 1, 2, 4, 8, and 16 shots. From the results, we observe that our CLIP-CITE performs very competitively, especially under 1, 2, and 4 shots. When compared with the second-best competitor MaPLe \cite{maple} on the average results, our CLIP-CITE demonstrates performance improvements by 3.42\%, 3.00\%, 2.48\%, 1.73\%, and 1.52\% in scenarios with 1, 2, 4, 8, and 16 shots, respectively. These gains underscore CLIP-CITE's effectiveness in generalizing to downstream tasks when provided with limited labeled examples. More comparisons of each dataset are provided in the supplementary materials.






\textbf{Results of BNG.} \Cref{tab:basetonew} showcases the BNG performance of our CLIP-CITE in comparison to five competing methods: CoOp \cite{coop}, CoCoOp \cite{cocoop}, MaPLe \cite{maple}, and CLIPood \cite{clipood}. The accuracy metrics are reported for both the base classes (\textbf{B}), new classes (\textbf{N}), and their harmonic mean (\textbf{HM}). From the results, we observe that our CLIP-CITE performs the best under both \textbf{B} and \textbf{N} metrics on the average of 11 datasets, leading to a notable 2.14\% improvement in the \textbf{HM} metric over the second-best competitor. In comparison to the original CLIP model without additional fine-tuning using the base data from the downstream task, our CLIP-CITE demonstrates a remarkable 16.14\% increase in base class accuracy across an average of 11 datasets. This indicates that fine-tuning CLIP with the base data could significantly improve the professionalism of CLIP in specific domains, which is also verified by the other competitors. Besides, our CLIP-CITE also obtains 3.86\% improvement over CLIP on the novel classes, which indicates that our CLIP-CITE improves the generalization capability under the open-vocabulary scenario. In contrast, while CoOp and CoCoOp also exhibit a notable enhancement in base accuracy, they compromise their capability for generalization to new classes. When compared with the competitors on the specific dataset, our CLIP-CITE performs the best on 9 out of 11 datasets in terms of \textbf{HM} metric. Moreover, we have noticed that CLIP-CITE showcases the exceptional performance, particularly on fine-grained datasets, with remarkable results observed on EuroSAT, Cars, Flowers102, and Aircrafts datasets. This leads us to speculate that fine-tuning can acquire finer and more specialized information.

From the BNG results in \Cref{tab:basetonew}, we could conclude that our method could effectively handle the overfitting and catastrophic forgetting issues as it could improve the performances on both base and novel classes from the same domain at the same time.


\begin{table}[!tb]
   \caption{Comparison performances (\%) on BNG task in terms of B, N, and HM metrics. The results of all the competitors are directly from the original literature. The best results are marked in \textbf{bold}. }
    \resizebox{\linewidth}{!}{
        \begin{tabular}{lc|@{\hspace{0.5cm}}c@{\hspace{0.5cm}}c@{\hspace{0.3cm}}c@{\hspace{0.3cm}}c@{\hspace{0.3cm}}c@{\hspace{0.5cm}}>{\columncolor{gray!20}}c}
        \toprule
        Method & & CLIP & CoOp & CoCoOp & MaPLe & CLIPood & \textbf{Ours} \\
        \midrule
        \multirow{3}{*}{Average on}              & B  & 69.34 & 82.69 & 80.47 & 82.28 & 83.90 & \textbf{85.48} \\
                                                 & N & 74.22 & 63.22 & 71.69 & 75.14 & 74.50 & \textbf{77.08} \\
                                                 & HM    & 71.70 & 71.66 & 75.83 & 78.55 & 78.92 & \textbf{81.06} \\
        \midrule
          
        \multirow{3}{*}{ImageNet}                & B  & 72.43 & 76.47 & 75.98 & 76.66 & 77.50 & \textbf{78.44} \\
                                                 & N & 68.14 & 67.88 & 70.43 & 70.54 & 70.30 & \textbf{71.07} \\
                                                 & HM    & 70.22 & 71.92 & 73.10 & 73.47 & 73.72 & \textbf{74.58} \\
        \midrule
        \multirow{3}{*}{Caltech101}              & B  & 96.84 & 98.00 & 97.96 & 97.74 & 98.70 & \textbf{98.82} \\
                                                 & N & 94.00 & 89.81 & 93.81 & 94.36 & \textbf{94.60} & 94.28 \\
                                                 & HM    & 95.40 & 93.73 & 95.84 & 96.02 & \textbf{96.61} & 96.50 \\
        \midrule
        \multirow{3}{*}{OxfordPets}              & B  & 91.17 & 93.67 & 95.20 & 95.43 & 95.70 & \textbf{96.01} \\
                                                 & N & 97.26 & 95.29 & 97.69 & 97.76 & 96.40 & \textbf{97.95} \\
                                                 & HM    & 94.12 & 94.47 & 96.43 & 96.58 & 96.05 & \textbf{96.97} \\
        \midrule
        \multirow{3}{*}{Cars}                    & B  & 63.37 & 78.12 & 70.49 & 72.94 & 78.60 & \textbf{82.83} \\
                                                 & N & 74.89 & 60.40 & 73.59 & 74.00 & 73.50 & \textbf{74.51} \\
                                                 & HM    & 68.65 & 68.13 & 72.01 & 73.47 & 75.96 & \textbf{78.45} \\
        \midrule
        \multirow{3}{*}{Flowers102}              & B  & 72.08 & \textbf{97.60} & 94.87 & 95.92 & 93.50 & 95.98 \\
                                                 & N & \textbf{77.80} & 59.67 & 71.75 & 72.46 & 74.50 & 76.45 \\
                                                 & HM    & 74.83 & 74.06 & 81.71 & 82.56 & 82.93 & \textbf{85.11} \\
        \midrule
        \multirow{3}{*}{Food101}                 & B  & 90.10 & 88.33 & 90.70 & 90.71 & 90.70 & \textbf{90.81} \\
                                                 & N & 91.22 & 82.26 & 91.29 & \textbf{92.05} & 91.70 & 91.55 \\
                                                 & HM    & 90.66 & 85.19 & 90.99 & 91.38 & \textbf{91.20} & 91.18  \\
        \midrule
        \multirow{3}{*}{Aircrafts}          & B  & 27.19 & 40.44 & 33.41 & 37.44 & 43.30 & \textbf{47.26} \\
                                                 & N & 36.29 & 22.30 & 23.71 & 35.61 & 37.20 & \textbf{38.37} \\
                                                 & HM    & 31.09 & 28.75 & 27.74 & 36.50 & 40.02 & \textbf{42.35} \\
        \midrule
        \multirow{3}{*}{SUN397}                  & B  & 69.36 & 80.60 & 79.74 & 80.82 & 81.00 & \textbf{82.30} \\
                                                 & N & 75.35 & 65.89 & 76.86 & 78.70 & 79.30 & \textbf{79.40} \\
                                                 & HM    & 72.23 & 72.51 & 78.27 & 79.75 & 80.14 & \textbf{80.82} \\
        \midrule
        \multirow{3}{*}{DTD}                     & B  & 53.24 & 79.44 & 77.01 & 80.36 & 80.80 & \textbf{84.26} \\
                                                 & N & 59.90 & 41.18 & 56.00 & 59.18 & 58.60 & \textbf{64.54} \\
                                                 & HM    & 56.37 & 54.24 & 64.85 & 68.16 & 67.93 & \textbf{73.09} \\
        \midrule
        \multirow{3}{*}{EuroSAT}                 & B  & 56.48 & 92.19 & 87.49 & 94.07 & \textbf{97.50} & 95.61 \\
                                                 & N & 64.05 & 54.74 & 60.04 & 73.23 & 64.10 & \textbf{80.59} \\
                                                 & HM    & 60.03 & 68.69 & 71.21 & 82.35 & 77.35 & \textbf{87.46} \\
        \midrule
        \multirow{3}{*}{UCF101}                  & B  & 70.53 & 84.69 & 82.33 & 83.00 & 85.70 & \textbf{87.56} \\
                                                 & N & 77.50 & 56.05 & 73.45 & 78.66 & \textbf{79.30} & 79.01 \\
                                                 & HM    & 73.85 & 67.46 & 77.64 & 80.77 & 82.38 & \textbf{83.07} \\
        \bottomrule    
        \end{tabular}   
    }
        \vspace{-0.2cm}

   \vspace{1.2mm}
   \label{tab:basetonew}
\end{table}

\begin{table}[!htb]
    \caption{\textbf{DG} performances (\%). All methods are trained on the ImageNet and evaluated on ImageNet-V2 (-V2), ImageNet-S (-S), ImageNet-A (-A), and ImageNet (-R).}
    \centering
    \resizebox{\linewidth}{!}{
        \begin{tabular}{lcccccc}
        \toprule
        \multirow{2}{*}{Method} & In-Distribution & \multicolumn{5}{c}{Out-of-Distribution} \\
        \cmidrule(lr){2-2}  \cmidrule(lr){3-7}
        & ImageNet & -V2 & -S & -A & -R  & Aver. \\
        \midrule
        Zero-shot & 66.7 & 60.8 & 46.1 & 47.8 & 74.8 & 57.2 \\
        Fine-tune & 68.2 & 61.9 & 46.8 & 46.4 & 75.1 & 57.6 \\
        CoOp & 71.5 & 64.2 & 48.0 & 49.7 & 75.2 & 59.3 \\
        CoCoOp & 71.0 & 64.2 & 48.8 & 50.6 & 76.2 & 59.9 \\
        MaPLe & 70.7 & 64.1 & 49.1 & \textbf{50.9} & 77.0 & 60.3 \\
        CLIPood & 71.6 & 64.9 & 49.3 & 50.4 & 77.2 & 60.4 \\
        \midrule
        CLIP-CITE (Ours) & \textbf{72.9} & \textbf{65.8} & \textbf{49.6} & 50.0 & \textbf{77.5} & \textbf{60.7} \\
        \bottomrule    
        \end{tabular}   
    }
    \vspace{10pt}
    \label{tab: domain shifts}
\end{table}

\begin{table*}
    \centering
\caption{Cross-domain generalization (\textbf{CDG}) evaluation (\%). All the models are trained on the base training set of 10 datasets and evaluated on the ImageNet dataset. Note that vanilla CLIP achieves 72.43\% and 68.14\% in terms of B and N metrics on ImageNet, respectively.}
    \resizebox{\linewidth}{!}{
        \begin{tabular}{l|cc|cc|cc|cc|cc|cc|cc|cc|cc|cc}
        \toprule
        \multirow{2}{*}{Method} & \multicolumn{2}{c|}{Caltech101} & \multicolumn{2}{c|}{OxfordPets} & \multicolumn{2}{c|}{Cars} & \multicolumn{2}{c|}{Flowers102} & \multicolumn{2}{c|}{Food101} & \multicolumn{2}{c|}{Aircrafts}  & \multicolumn{2}{c|}{SUN397} & \multicolumn{2}{c|}{DTD} & \multicolumn{2}{c|}{EuroSAT} & \multicolumn{2}{c}{UCF101}\\
        \cmidrule{2-21}
        & B & N & B & N & B & N & B & N & B & N & B & N & B & N & B & N & B & N & B & N\\
        \midrule
        CoOP \cite{coop} & 55.72 & 54.62 & 58.20 & 50.38 & 53.72 & 50.91 & 46.88&37.43 &45.53&42.51 &62.02&56.15 &52.52&54.59 & 56.94&55.87 & 51.81&49.69 & 42.64&39.76 \\
        CoCoOp \cite{cocoop} & 70.65 & 66.97 &68.80 &59.75 & 58.54 & 56.89 &50.91&55.78 &66.08&62.89 &55.81&56.76 &69.57&67.28 &65.26&62.64 &60.51&58.66 &51.42&54.64\\
        MaPLe \cite{maple} & 72.48 & 69.54 &63.91 &48.84  & 68.65 & 66.03 & 60.14 & 47.40  & 71.84&67.78 & 67.44 &62.66 & 70.05&67.83 &65.56&63.98 &40.43&45.80 & 64.27&62.28\\
        \midrule
         CLIP-CITE & 72.24 & \textbf{69.57} & \textbf{72.38} & \textbf{68.44} & \textbf{72.30} & \textbf{69.04} & \textbf{72.06} & \textbf{68.68} & \textbf{72.87} & \textbf{69.03} & \textbf{72.54} & \textbf{68.49} & \textbf{72.33} & \textbf{68.89} & \textbf{71.95} & \textbf{68.44} & \textbf{72.29} & \textbf{68.38} & \textbf{72.81} & \textbf{68.59} \\ 
        \bottomrule        
        \end{tabular}
    }
    \vspace{10pt}
    \label{tab:to-imagenet}
\end{table*}

\textbf{Results of DG.}
The DG performances of our method, along with six competitors, are presented in \Cref{tab: domain shifts}. In this evaluation, the model is trained on the few-shot ImageNet dataset and then tested on different datasets, namely ImageNetv2, ImageNet-Sketch, ImageNet-A, and ImageNet-R, which have the same class labels as ImageNet but belong to different domains. Our method demonstrates superior performance in terms of in-domain ImageNet accuracy, achieving an accuracy of 72.9\%. Additionally, our method achieves a high average accuracy of 60.7\% across the out-of-domain datasets, surpassing all existing methods except for ImageNet-A. These results indicate that our method is effective in handling domain shifts. 


\textbf{Results of CDG.} The CDG comparison results of our method and three competitors are displayed in \Cref{tab:to-imagenet}. Specifically, we fine-tune the model with the training data from various datasets and then evaluate the model on the test data of the ImageNet dataset.  For ease of comparison with the results presented in \Cref{tab:basetonew}, we report \textbf{B} and \textbf{N} performance metrics on the ImageNet dataset. From the results, we observe that our CLIP-CITE could maintain its performance on the ImageNet dataset regardless of the datasets used for training, indicating the robustness of the proposed method. In contrast, the performances of the other competitors on ImageNet drop significantly. When considering the results presented in \Cref{tab:basetonew}, we observe that the existing competitors that tuning the CLIP model using a specific dataset, their performance on that dataset notably improves, particularly in terms of the \textbf{B} metric. However, their performance on other datasets significantly declines. This suggests that current competitors enhance their professionalism when fine-tuned with a specific dataset but at the cost of losing their versatility, an issue known as catastrophic forgetting. In contrast, our fine-tuning strategy not only enhances the CLIP's professionalism but also maintains its versatility. Furthermore, we observed that as the domain difference between the training dataset and the ImageNet dataset increases, the performance of the model on the ImageNet dataset degrades more severely. For example, the \textbf{B} performance on ImageNet falls from 72.43\% to 40.43\% when fine-tuning the model with Maple \cite{maple} on the EuroSAT dataset. We postulate that this is primarily because the learnable parameters of the parameter-efficient competitors primarily capture domain- and class-specific information, making them less suitable for novel classes from different domains. In contrast, our fully fine-tuning method distributes the changes in domain and category equally across the parameters of the model, resulting in small changes in parameter magnitude, which enables it to effectively handle different domains and categories simultaneously. Furthermore, our distillation strategy also benefits in  mitigating catastrophic forgetting.

\subsection{Further Analysis}

\textbf{Effects of different objectives.}
\cref{tab: different objectives} displays the ablation study of our CLIP-CITE with various training objectives on the BNG task of ImageNet. The first row represents the results obtained with the basic CLIP model. When fine-tuning the model with only 
$\mathcal{L}_{DVA}$, it achieves a 2.78\% improvement in \textbf{HM} compared to the naive CLIP. Additionally, the introduction of of supervised contrastive learning objective $\mathcal{L}_{SCL}$ leads to further improvement in both \textbf{B} and \textbf{N} metrics. By combining both objectives ($\mathcal{L}_{DVA}$ + $\mathcal{L}_{SCL}$), the performance of both \textbf{B} and \textbf{H} metrics continue to improve. Furthermore, incorporating the vision-language similarity distillation loss $\mathcal{L}_{VLD}$ into the objective results in the best performance of 74.58\% \textbf{HM} accuracy. These experimental outcomes highlight the efficacy of each objective function introduced in this work.

 \begin{figure}[t]
    \centering
    \includegraphics[width=\linewidth]{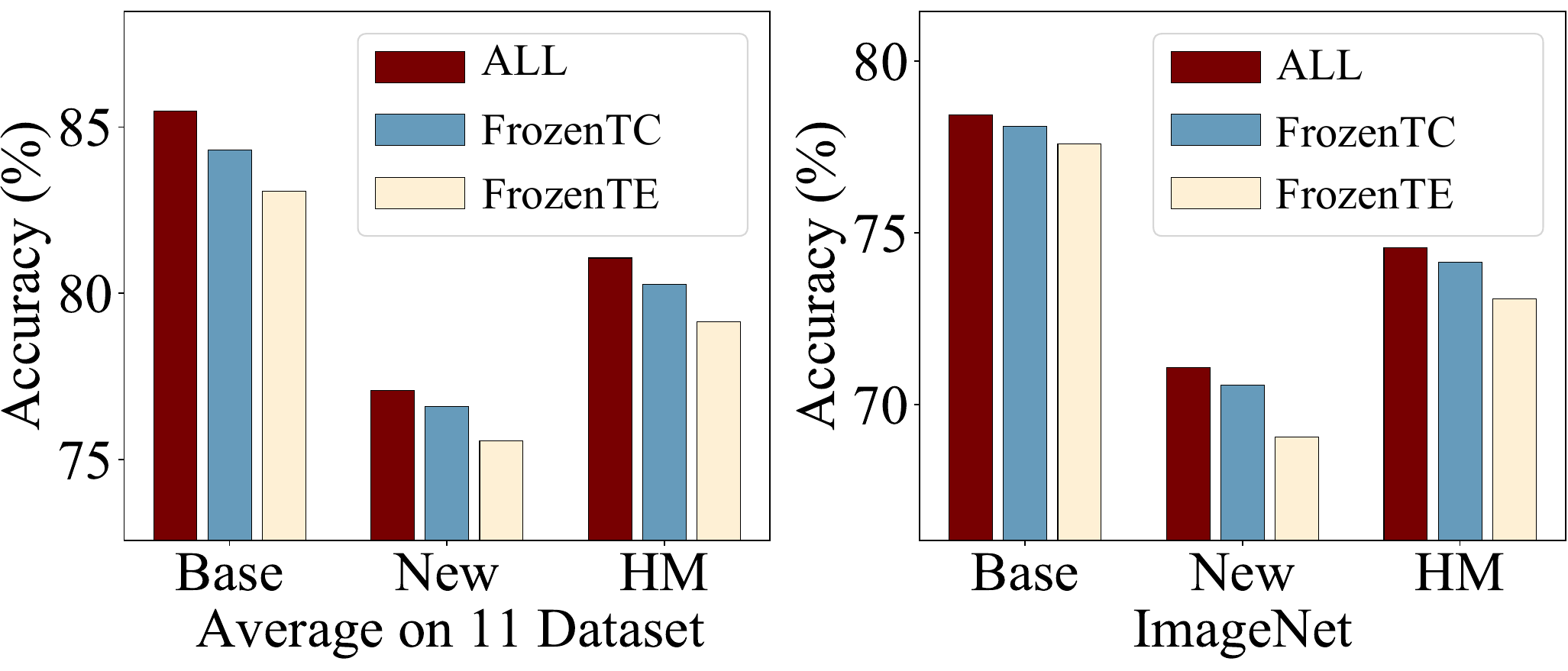} 
    \caption{Ablation on different fine-tuning parts of the model.}
    \vspace{15pt}
    \label{fig:ft-parts}
\end{figure}

\begin{table}[t]
    \centering
        \caption{Ablation results (\%) of our CLIP-CITE with various training objectives on the BNG task of the ImageNet dataset.}
    \vspace{5pt}
    \resizebox{0.98\linewidth}{!}{
        \begin{tabular}{c@{\hspace{0.8cm}}c@{\hspace{0.8cm}}c@{\hspace{0.3cm}}|@{\hspace{0.3cm}}c@{\hspace{0.8cm}}c@{\hspace{0.8cm}}c}
        \toprule
          $\mathcal{L}_{DVA}$ & $\mathcal{L}_{SCL}$ & $\mathcal{L}_{VLD}$ & \footnotesize{B} & \footnotesize{N} & \footnotesize{HM }\\
        \midrule
          &  & & 72.43 & 68.14 & 70.22 \\
          \checkmark &  &  & 77.35 & 69.12 & 73.00 \\
          & \checkmark & & 78.10 & 70.67 & 74.20 \\
          \checkmark & \checkmark &  & \textbf{78.49} & 70.76 & 74.43 \\
         \checkmark &  & \checkmark & 77.31 & 70.20 & 73.58 \\
          \checkmark & \checkmark & \checkmark & 78.44 & \textbf{71.07} & \textbf{74.58} \\
        \bottomrule
        \end{tabular}
    }
    \vspace{15pt}
    \label{tab: different objectives}
\end{table}

\begin{figure}[t]
    \centering
    \vspace{10pt}
    \includegraphics[width=2.8in,height=1.8in]{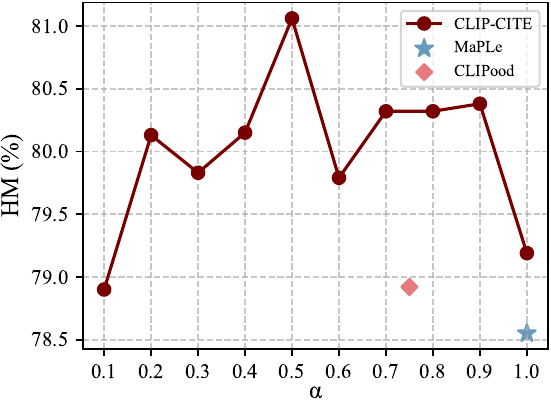}
    \caption{Comparison results with the different ensemble ratio $\alpha$.}
    \vspace{20pt}
    \label{fig:ensemble}
\end{figure}

\begin{table}[!htb]
    \centering
        \caption{Comparison performances (\%) and training efficiency of the existing prompt learning methods and ours. All the models are trained on a single NVIDIA GeForce RTX 3090 GPU.}
    \resizebox{\linewidth}{!}{
        \begin{tabular}{l|c|ccc|cc}
        \toprule  
        \multirow{2}{*}{Method} & \multirow{2}{*}{Iterations}  & \multicolumn{3}{c|}{ImageNet} & \multicolumn{2}{c}{Training Resources} \\
        \cmidrule{3-5} \cmidrule{6-7}
        & & Base & New & HM & Training-time & GPU-usage\\
        \midrule
        CLIP \cite{clip} & N/A & 72.43 & 68.14 & 70.22 & $N/A$ & $N/A$ \\
        CoOp \cite{coop} & 12.5~K & 76.47 & 67.88 & 71.92 & $\approx1~h$ &  $\approx10~G$ \\
        CoCoOp \cite{cocoop} & 80K & 75.98 & 70.43 & 73.10 &  $>7~h$ & $\approx10~G$ \\
        MaPLe \cite{maple} & 10~K  & 76.66 & 70.54 & 73.47 & $\approx45~min$ &  $\approx10~G$\\
        \midrule 
        CLIP-CITE & 1.2K & \textbf{78.44} & \textbf{71.07} & \textbf{74.58} &  $\approx20~min$ &  $\approx19~G$\\
        \bottomrule
        \end{tabular}    
    }
    \vspace{10pt}

    \label{tab: efficiency}
\end{table}

\textbf{Effects of the fine-tuning parts.} In this experiment, we conduct an ablation study to examine the effects of different fine-tuning parts. The average results of 11 datasets and the results on ImageNet dataset are shown in \cref{fig:ft-parts}. \textbf{FrozenTC} indicates that the text embeddings are taken as the classifiers of the visual feature representations and are frozen during optimizing \cref{eq:dva}. \textbf{FrozenTE} indicates that the text encoder is frozen during optimizing \cref{eq:itm}. \textbf{ALL} indicates that all the parameters of the model are fine-tuning during training. From the results in \cref{fig:ft-parts}, we observe that HM performance of \textbf{ALL} witnesses a considerable lift compared with those of \textbf{FrozenTC} and \textbf{FrozenTE}, which concludes that comprehensive fine-tuning enhances model capabilities more effectively than partial fine-tuning.

\textbf{Effects of the weight ensemble.} We investigate the effect of weight ensemble in \cref{fig:ensemble}. The results therein lead us to the conclusion that even without the weight ensemble inference (with $\alpha$ set to 1.0), our method still delivers noteworthy performance with results of 85.79\% (B), 73.52\% (N), and 79.19\% (HM) on the BNG task. Notably, it outperforms CLIPood, which integrates model weight inference ensemble, and MaPLe, by achieving a lift of 0.27\% and 0.64\% in the {HM} metric, respectively. Moreover, with the appropriate weight ensemble ratio (setting $\alpha$ to 0.5), we have noticed a notable improvement in both base and novel performance.

\textbf{Training efficiency.} \cref{tab: efficiency} presents a comprehensive comparison of our CLIP-CITE and four parameter-efficient competitors. \textit{The results indicate that parameter efficiency does not necessarily translate to computational efficiency.} Specifically, our model, despite fine-tuning more parameters and utilizing more GPU resources, demonstrates superior performance with significantly fewer training iterations and shorter overall training time compared to the parameter-efficient competitors. Although the prompt-based methods offer parameter efficiency, they still necessitate the backpropagation of the entire model, along with numerous training iterations, to achieve convergence.

\begin{figure}[!tb]
    \centering
    \includegraphics[width=\linewidth]{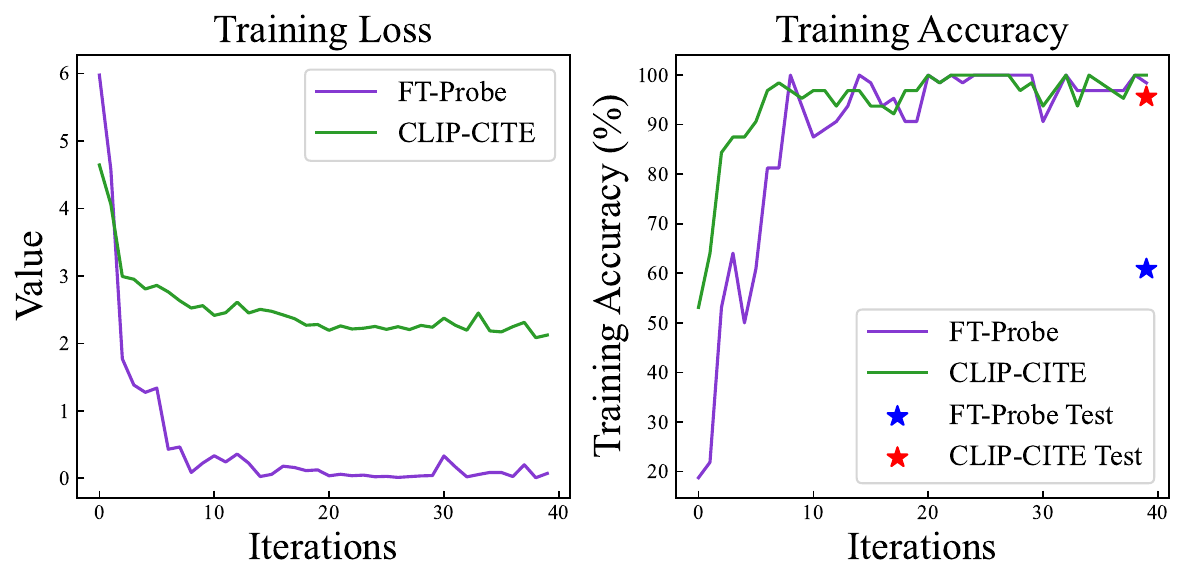}
    \caption{Training loss and accuracy of FT-Probe and CLIP-CITE on the EuroSAT dataset.}
    \vspace{30pt}
    \label{fig:loss}
\end{figure}

\textbf{Overfitting analysis.} As shown in \cref{fig:loss}, the FT-Probe model exhibits a gradual decrease in training loss and a corresponding increase in accuracy on the training set. However, the final test set accuracy is only 60.86\% (the blue star), indicating the presence of overfitting. Conversely, our CLIP-CITE model also demonstrates a reduction in the loss function and a consistent improvement in training set accuracy. Notably, it achieves a significantly higher test set accuracy of 95.61\%, suggesting that our approach effectively overcomes the issue of overfitting. This underscores the importance of addressing overfitting when fully fine-tuning models and demonstrates the effectiveness of our CLIP-CITE method.

\textbf{Prompt Learning with proposed loss.} To evaluate the effectiveness of full-fine-tuning, we also explore the prompt learning methods with our proposed loss. The results, detailed in \Cref{tab: PL+different objectives}, indicate that prompt learning methods experience a modest improvement with the implementation of our proposed loss functions \textit{i.e.} $L_{SCL}$ and $L_{VLD}$. Notably, our CLIP-CITE still maintains a performance edge. Besides, with the simple fine-tuning (FT-Probe), the tuned model seems to be overfitting, as shown in \cref{fig:ft-methods}. Therefore, we propose that both full fine-tuning and well-designed loss functions are crucial in adapting VLMs to the downstream few-shot tasks.

\begin{table}[!h]
    \centering
    \resizebox{0.98\linewidth}{!}{
        \begin{tabular}{l@{\hspace{0.8cm}}c@{\hspace{0.8cm}}c@{\hspace{0.3cm}}|@{\hspace{0.3cm}}c@{\hspace{0.8cm}}c@{\hspace{0.8cm}}c}
        \toprule
        Method &  $\mathcal{L}_{SCL}$ & $\mathcal{L}_{VLD}$ & \footnotesize{B} & \footnotesize{N} & \footnotesize{HM }\\
        \midrule
            CLIP  & & & 72.43 & 68.14 & 70.22 \\
        \midrule
             CoOp & & & 76.47 & 67.88 & 71.92 \\
             CoOp & \checkmark & & 76.51 & 67.93 & 71.97 \\
             CoOp & \checkmark & \checkmark & 78.23 & 70.89 & 72.11 \\
        \midrule
             MaPLe & & & 76.66 & 70.54 & 73.47 \\
             MaPLe & \checkmark & & 76.70 & 70.67 & 73.56 \\
             MaPLe & \checkmark & \checkmark & 76.71 & 70.89 & 73.69 \\ 
        \midrule
             CLIP-CITE & \checkmark & \checkmark & 78.44 & 71.07 & 74.58 \\
        \bottomrule
        \end{tabular}
    }
    \vspace{10pt}
    \caption{Ablation results (\%) of our CLIP-CITE and prompt learning with various training objectives on the BNG task of the ImageNet dataset.}
    \vspace{15pt}
    \label{tab: PL+different objectives}
\end{table}

\begin{figure}[!htb]
    \centering
    \scriptsize
   \begin{overpic}[width=\linewidth,height=0.4\linewidth]{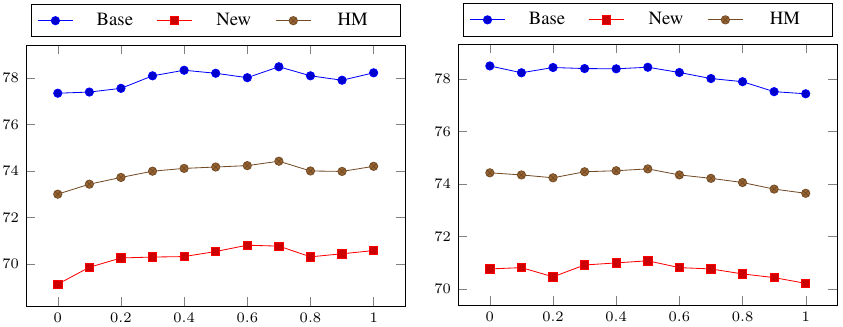} 
   \scriptsize
   \put(15,-5){(a)~Impacts of $\lambda$.}
   \put(65,-5){(b)~Impacts of $\eta$.}
   \end{overpic}   
   \vspace{10pt}
    \caption{Impacts (\%) of the hyper-parameter $\lambda$ and $\eta$ on the BNG performances. We report the results on the ImageNet dataset.}
    \vspace{20pt}
    \label{fig:hyperparameter}
\end{figure}

\textbf{The effect of the hyper-parameter $\lambda$ and $\eta$.} In \cref{fig:hyperparameter}, we ablate the different values on $\lambda$ and $\eta$ in \cref{eq:total-loss}. From the results, we observe that the performances in terms of HM are better when applying the $\mathcal{L}_{SCL}$, e.g., $\lambda$ is greater than 0. It indicates that supervised vision-language alignment is necessary when fine-tuning. Besides, the vision-language similarity distillation can regularize the model well when $\eta$ is less than 0.1. In the experiments, the optical $\lambda$ and $\eta$ are set to 0.7 and 0.1, respectively.

\begin{figure}[!h]
   \begin{overpic}[width=\linewidth]{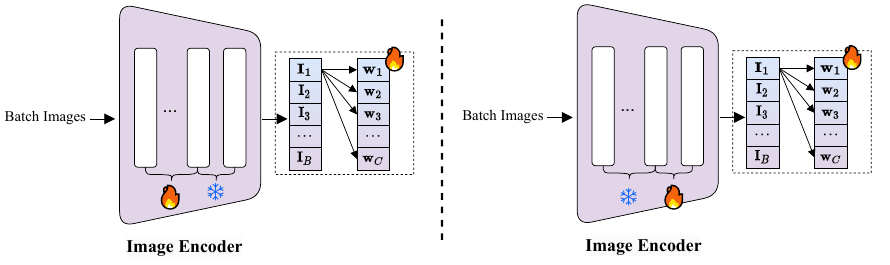} 
   \vspace{10mm}
   \scriptsize
   \put(6, -4){(a)~Fine-tuning previous layers.}
   \put(58, -4){(b)~Fine-tuning late layers.}
   \end{overpic}
    \vspace{5pt}
   \caption{Illustration of fine-tuned model within the distinct layers. (a) illustrates layers preceding the image encoder, while (b) delineates layers succeeding the image encoder.}
   \label{fig:ft-ratio}
   \vspace{20pt}
\end{figure}

\begin{figure}[!h]
   \begin{overpic}[height=3.5cm, width=\linewidth]{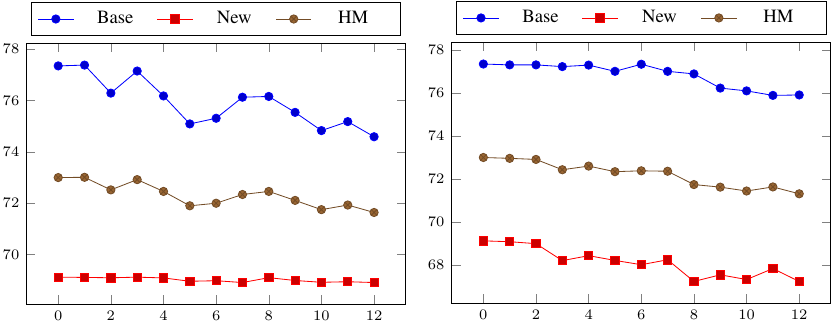} 
   \scriptsize
      \vspace{10mm}
   \put(6, -4){(a)~Freezing the late $i_{th}$ layers.}
   \put(58, -4){(b)~Freezing the previous $i_{th}$ layers.}
   \end{overpic}
\vspace{10pt}
    \caption{The effect of the fine-tuning layers. (a) indicates we fine-tune the previous layers and freeze the $i_{th}$  late layers corresponding to \cref{fig:ft-ratio}. (a), while (b) indicates we freeze the previous $i_{th}$ layers and fine-tune the late layers corresponding to \cref{fig:ft-ratio}. (b). }
   \label{fig:frozen-layer}
   \vspace{20pt}
\end{figure}

\textbf{The effect of the full fine-tuning.} \cref{fig:ft-ratio} shows the different fine-tuning manners of the image encoder, \textit{e.g.} fine-tuning previous layers and fine-tuning late layers. And we conduct the experiments with $\mathcal{L}_{DVA}$ for ablation. \cref{fig:frozen-layer}. (a) shows the results that we fine-tune previous layers and freeze the late layers, while \cref{fig:frozen-layer}. (b) the results that we fine-tune previous layers and freeze the late layers. From the experimental results, we observe that when there are only a few frozen layers, the performance is comparable to full fine-tuning. However, as the number of frozen layers increases, the effectiveness diminishes, \textit{i.e.} the last 3 frozen layers led to a decline in the results shown in \cref{fig:frozen-layer}. (a). Overall, full fine-tuning is better than partial fine-tuning.

%% file: tex/conclusion.tex
\section{Conclusion}
In this paper, we have presented CLIP-CITE, a fine-tuning approach designed to adapt CLIP for downstream tasks in limited-data scenarios. By devising a discriminative visual-text alignment task, implementing supervised contrastive loss, and employing visual-language similarity distillation, CLIP-CITE effectively addresses the common issues of overfitting and catastrophic forgetting encountered by existing fine-tuning methods. Our experimental results demonstrate that a carefully crafted fine-tuning strategy can enable CLIP to acquire both domain-specific and class-specific knowledge, while maintaining its versatility across other domains and classes. Notably, despite involving the tuning of more parameters, our approach offers superior computational efficiency compared to parameter-efficient prompt-based competitors.